\documentclass[runningheads]{llncs}
\usepackage{graphicx}
\usepackage{amsmath}
\usepackage{subfigure}
\usepackage{algorithm}
\usepackage{algorithmic}
\usepackage{multirow}
\usepackage{threeparttable}
\usepackage{booktabs}
\usepackage{amssymb}
\usepackage[pagebackref=true,breaklinks=true,letterpaper=true,colorlinks,bookmarks=false]{hyperref}

\usepackage{enumitem}
\setitemize{noitemsep,topsep=0pt,parsep=0pt,partopsep=0pt}

\usepackage{color}

\begin{document}
	\title{Cross-View Image Synthesis with Deformable Convolution and Attention Mechanism 
	}

	\titlerunning{Cross-View Image Synthesis}
	%
	\author{Hao Ding\inst{1} \and Songsong Wu\inst{1} \and
	Hao Tang\inst{2} \and Fei Wu\inst{1} \and Guangwei Gao\inst{1} \and Xiao-Yuan Jing\inst{3} }
	
	\authorrunning{H. Ding et al.}
	%
	\institute{College of Automation and College of Atificial Intelligence, Nanjing University of Posts and Telecommunications, Nanjing 210023, China\\ \email{1981716375@qq.com,sswuai@126.com,wufei\_8888@njupt.edu.cn,csggao@gmail.com}\\
	\and University of Trento    \email{hao.tang@unitn.it}\\
	\and Wuhan University        \email{Jingxy\_2000@126.com}
	}

	\maketitle              

	\begin{abstract}
		Learning to generate natural scenes has always been a daunting task in computer vision. This is even more laborious when generating images with very different views. When the views are very different, the view fields have little overlap or objects are occluded, leading the task very challenging. In this paper, we propose to use Generative Adversarial Networks (GANs) based on a deformable convolution and attention mechanism to solve the problem of cross-view image synthesis  (see Fig.~\ref{fig0}).
		It is difficult to understand and transform scenes appearance and semantic information from another view,
		thus we use deformed convolution in the U-net network to improve the network's ability to extract features of objects at different scales.
		Moreover, to better learn the correspondence between images from different views, we apply an attention mechanism to refine the intermediate feature map thus generating more realistic images.
		A large number of experiments on different size images on the Dayton dataset~\cite{1} show that our model can produce better results than state-of-the-art methods.
		
	\end{abstract}
	\section{Introduction}
\begin{figure}[!t]
	\centerline{\includegraphics[width=0.8\linewidth]{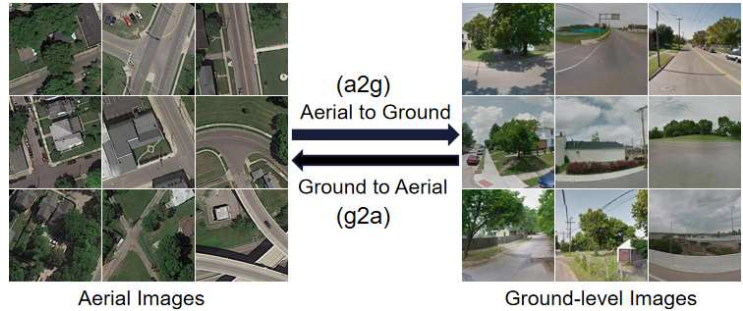}}
	\caption{Example images in overhead/aerial view (left) and street-view/ground-level (right). The images reflect the great diversity and richness of features in two views implying that the network needs to learn a lot for meaningful cross-view generation.}
	\label{fig0}
\end{figure}

	 Cross-view image synthesis aims to translate images between two distinct views, such as synthesizing ground images from aerial images, and vice versa. This problem has aroused great interest in the computer vision and virtual reality communities, and it has been widely studied in recent years~\cite{2,3,4,5,6,7,8,9}. Earlier work used encoder-decoder convolutional neural networks (CNNs) to study the viewpoint code included in the bottleneck representation for urban scene synthesis~\cite{10} and 3D object transformations~\cite{11}. Besides, when the view fields have little overlap or objects are occluded, 
	 and similar objects in one view may be completely different from another view (i.e., view invariance issues), this task will be more challenging. For example, the aerial view of a building (i.e., the roof) tells very little about the color and design of the building seen from the street-view. The generation process is generally easier when the image contains a single object in a uniform background. In contrast, when the scene contains multiple objects, generating other view becomes much more challenging. This is due to the increase in underlying parameters that contribute to the variations (e.g., occlusions, shadows, etc). An example scenario, addressed here, is generating street-view (a.k.a ground level) image of a location from its aerial (a.k.a overhead) image. Fig.~\ref{fig0} illustrates some corresponding images in the two different views.
	 
 To solve this challenging problem, Krishna and Ali~\cite{6} proposed a conditional GAN model that jointly learns the generation in both the image domain and the corresponding semantic domain, and the semantic predictions are further utilized to supervise the image generation. Although this method has been interestingly explored, there are still unsatisfactory aspects of the generated scene structure and details.
 Moreover, Tang et al.~\cite{12} recently proposed the multi-channel attention selection generation adversarial network (SelectionGAN), which can learn conditional images and target semantic maps together, and the automatically learned uncertainty map can be used to guide pixel loss to achieve better network optimization. However, we observe that there are still unsatisfactory aspects in the generated scene structure and details. 
For example, for the outline boundaries of some objects, there are obvious wrong marks and unclear.
	
To tackle this challenging problem, we add deformed convolution to the U-net network to improve the network's ability to extract features of objects at different scales. At the same time, we use the attention mechanism~\cite{13} to refine the feature map to obtain a more detailed feature map for generating more realistic images. A large number of experiments show that our model can produce better results than state-of-the-art models, i.e., Pix2Pix~\cite{2}, X-Fork~\cite{6}, X-Seq~\cite{6} and SelectionGAN~\cite{12}.
	
In summary, our contributions of this paper are as follows:
\begin{itemize}
	\item We employed the attention mechanism to refine the feature map to generate more realistic images for the challenging cross-view image translation tasks.
	\item We also embed deformable convolutions in the U-net network to improve the network's ability for extracting features of objects at different scales.
	\item An additional loss function is added to improve the network training, thereby achieving a more stable optimization process.
\end{itemize}

\section{Related work}
    Existing work on viewpoint transformation has been performed to synthesize novel views of the same object~\cite{14,15,16}. 
    For example, Zhou et al.~\cite{16} proposed models learn to copy pixel information from the input view and uses them to retain the identity and structure of the object to generate a new view. Tatarchenko et al.~\cite{15} trained a network of codes to obtain 3D representation models of cars and chairs, which were subsequently used to generate different views of unseen images of cars or chairs. Dosovitskiy et al.~\cite{14} learned to generate models by training 3D renderings of cars, chairs, and tables, and synthesize intermediate views and objects by interpolating between views and models. Zhai et al.~\cite{17} explored the semantic layout of predicting ground images from their corresponding aerial images. They synthesized ground panoramas using the predicted layouts. Previous work on aerial and ground images has addressed issues such as cross-view co-localization~\cite{18,19}, ground-to-aerial geo-localization~\cite{20} and geo-tagging the cross-view images~\cite{21}.
    
	
Compared with existing methods such as Restricted Boltzmann Machines~\cite{22} and Deep Boltzmann Machines~\cite{23}, generative adversarial networks (GANs)~\cite{24} have shown the ability to generate better quality images~\cite{25,26,27,28}. The vanilla GAN model~\cite{24} has two important components, i.e., the generator $G$ and the discriminator $D$.
The generator $G$ aims to generate realistic from the noise vector, while $D$ tries to distinguish between real image and image generated by $G$. Although it has been successfully used to generate high visual fidelity images~\cite{26,29,30,31}, there are still some challenges such as how to control the image generation process under specific settings. To generate domain-specific images, the conditional GAN (CGAN)~\cite{28} has been proposed. CGAN usually combines vanilla GAN with some external information.
	
Krishna and Ali~\cite{6} proposed two structures (X-Fork and X-Seq) based on Conditional GANs to solve the task of image translation from aerial to street-view using additional semantic segmentation maps. Moreover, Tang et al.~\cite{12} proposed the multi-channel attention selection generation adversarial network (SelectionGAN), which consists of two generation stages. In the first stage, a cyclic semantically guided generation sub-net was proposed. 
This network receives images and conditional semantic maps in one view, while synthesizing images and semantic maps in another view. The second stage uses the rough predictions and learned deep semantic features of the first stage, and uses the suggested multi-channel attention selection the module performs fine-grained generation.
		
\section{Network Design}
\begin{figure}[!t]
	\centerline{\includegraphics[width=0.8\linewidth]{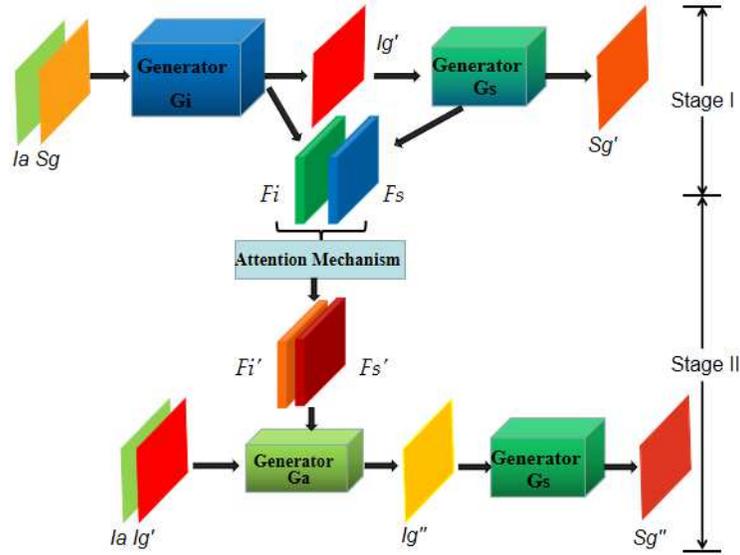}}
	\caption{Architecture of the proposed network.}
	\label{fig2}
\end{figure}

The network structure we proposed is based on the SelectionGAN model, which consists of three generators (i.e., $G_i$, $G_a$, $G_s$), two discriminators (i.e., $D_1$, $D_2$), and an attention mechanism module. The network structure can be divided into two stages, as shown in Fig.~\ref{fig2}. 

In the first stage, an image $I_{a}$ of one perspective and a semantic map $S_{g}$ of another perspective are input to the generator $G_i$ to generate an image $I_{g}^{\prime}$ of another perspective and the feature map $F_{i}$ of the last convolution layer. 
Then the generated image $I_{g}^{\prime}$ is input into the generator $G_s$ to generate the corresponding semantic map $S_{g}^{\prime}$.

In the second stage, the feature maps $F_{i}$ and $F_{s}$ generated in the first stage are refined through the attention mechanism module to obtain the refined feature maps $F_{i}^{\prime}$ and $F_{s}^{\prime}$. 
Next, they are combined with the image $I_{a}$ and the generated image $I_{g}^{\prime}$ and inputted to the generator $G_a$ to generate a refined image $I_{g}^{\prime \prime}$ as the final output.
This refined image $I_{g}^{\prime \prime}$ is then input to the generator $G_s$ to generate the corresponding semantic map $S_{g}^{\prime \prime}$. 

Note that we use only one generator $G_s$ in both the first and second stages, since the purpose is to generate a corresponding semantic image from an image.

\subsection{Attention Mechanism}
\label{ssec:subhead}
\begin{figure}[t]
	\centerline{\includegraphics[width=0.8\linewidth]{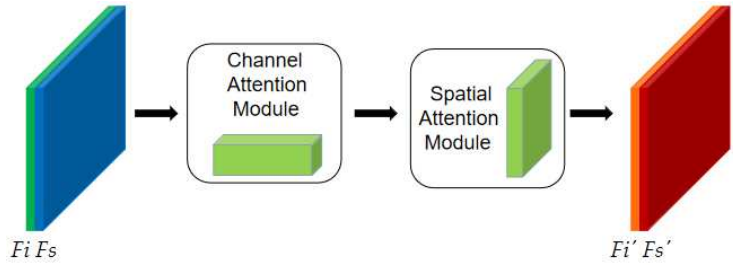}}
	\caption{Attention Mechanism Module.}
	\label{fig3}
\end{figure}

Since the SelectionGAN model takes the coarse feature map as input of the second stage. So we consider that we can use the attention mechanism to refine the feature map before inputting it into the generator $G_a$. 
The attention mechanism is consisted of Channel Attention Module and Spatial Attention Module, as shown in Fig.~\ref{fig3}. Given an intermediate feature map, the attention mechanism will follow two separate dimensions to infer the attention maps, and then the attention maps are multiplied with the input features to map adaptive features. Experiments show that after adding the attention mechanism, the generation performance is indeed improved.
	
\subsection{Deformable Convolution}
\label{ssec:subhead}
Deformable convolution~\cite{32} adds spatial sampling positions with additional offsets and learns offsets in the target task without additional supervision. The new module can easily replace the ordinary peers in existing CNNs and a large number of experiments have verified that this method learns dense spatial transformations in deep CNNs and is effective for complex visual tasks such as object detection and semantic segmentation.

\begin{figure}[t]
	\centerline{\includegraphics[width=0.8\linewidth]{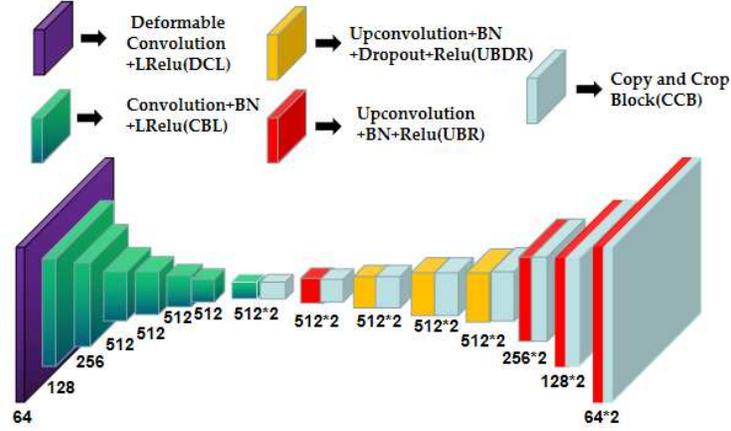}}
	\caption{Network structure of the proposed generator. BN means batch-normalization layer.}
	\label{fig4}
\end{figure}

Therefore, we embed deformable convolutions into U-net. The outermost layer of the network can better extract the features from the input maps. The network structure is shown in Fig.~\ref{fig4}.
	
\subsection{Overall Optimization Objective}
\noindent \textbf{Adversarial Loss.} SelectionGAN~\cite{12} uses one discriminator $D_1$ for the generated images on two stages.
$D_1$ takes the input and the generated fake image as input, however, the semantic map is not take into consideration.
Therefore, we propose a new discriminator $D_2$, which  also takes the semantic map  as input. The proposed semantic-guided adversarial losses can be expressed as follows,
	
\begin{equation}
\begin{aligned}
& \mathcal{L}_{cGAN}\left(I_{a}\textcircled{+}S_{g}, I_{g}^{\prime}\textcircled{+} S_{g}^{\prime}\right) \\= &
\mathbb{E}
\left[\log D_2\left(I_{a}\textcircled{+} S_{g}, I_{g}\textcircled{+} S_{g}\right)\right] \\
+ & \mathbb{E}
\left[\log \left(1-D_2\left(I_{a}\textcircled{+} S_{g}, I_{g}^{\prime}\textcircled{+} S_{g}^{\prime}\right)\right)\right],
\end{aligned}
\label{eq:1}
\end{equation}
\begin{equation}
\begin{aligned}
& \mathcal{L}_{cGAN}\left(I_{a}\textcircled{+} S_{g}, I_{g}^{\prime \prime}\textcircled{+} S_{g}^{\prime \prime}\right)\\ = &
\mathbb{E}
\left[\log D_2\left(I_{a}\textcircled{+} S_{g}, I_{g}\textcircled{+} S_{g}\right)\right]\\
+ & \mathbb{E}
\left[\log \left(1-D_2\left(I_{a}\textcircled{+}S_{g}, I_{g}^{\prime \prime}\textcircled{+}S_{g}^{\prime \prime}\right)\right)\right],
\end{aligned}
\label{eq:2}
\end{equation}
where the symbol $\textcircled{+}$ denotes the channel-wise concatenation operation.
Thus, the total adversarial loss can be formulated as follows,

\begin{equation}
\begin{aligned}
\mathcal{L}_{cGAN}= & \mathcal{L}_{cGAN}\left(I_{a}, I_{g}^{\prime}\right)+\lambda \mathcal{L}_{c G A N}\left(I_{a}, I_{g}^{\prime \prime}\right) \\
+ & \mathcal{L}_{c G A N}\left(I_{a}\textcircled{+} S_{g}, I_{g}^{\prime} \textcircled{+} S_{g}^{\prime}\right) \\
+ & \lambda \mathcal{L}_{cGAN}\left(I_{a}\textcircled{+}S_{g}, I_{g}^{\prime \prime}\textcircled{+}S_{g}^{\prime \prime}\right),
\end{aligned}
\label{eq:adv_loss}
\end{equation}
where $\mathcal{L}_{cGAN}\left(I_{a}, I_{g}^{\prime}\right)$ and $\mathcal{L}_{c G A N}\left(I_{a}, I_{g}^{\prime \prime}\right)$ are the adversarial losses defined in SelectionGAN.
	
\noindent \textbf{Overall Loss.} The total optimization loss is a weighted sum of several losses. The generators $G_i$, $G_s$, $G_a$ and discriminators $D_1$, $D_2$ are trained in an end-to-end fashion optimizing the following min-max function,
\begin{equation}
\setlength{\abovedisplayskip}{3pt}
\setlength{\belowdisplayskip}{3pt}
\min _{\left\{G_{i}, G_{s}, G_{a}\right\}} \underset{\{D_1, D_2\}}{\max} \mathcal{L}=\sum_{i=1}^{4} \lambda_{i} \mathcal{L}_{p}^{i}+\mathcal{L}_{c G A N}+\lambda_{t v} \mathcal{L}_{t v},
\label{eq:loss}
\end{equation}
where $\mathcal{L}_{p}^{i}$ uses the $L1$ reconstruction to separately calculate the pixel loss between the generated images $I_{g}^{\prime}$, $S_{g}^{\prime}$, $I_{g}^{\prime \prime}$ and $S_{g}^{\prime \prime}$ and the corresponding real ones. $\mathcal{L}_{tv}$ is the total variation regularization on the final synthesized image $I_{g}^{\prime \prime}$. $\lambda_{i}$ and $\lambda_{tv}$ are the trade-off parameters to control the relative importance of different objectives.

	\section{Experiments}
	
\noindent \textbf{Datasets.}
We follow~\cite{6,12,33} and perform extensive experiments on the challenging Dayton dataset in a2g (aerial-to-ground) and g2a (ground-to-aerial) directions with two different image resolutions (i.e., $256 {\times} 256$ and $64 {\times} 64$).
Specifically, we select 76,048 images and create a train/test split of 55,000/21,048 pairs. The images in the original dataset have $354 {\times} 354$ resolution.
We then resize them to $256 {\times} 256$.
	
	\begin{table}[!t]
	\centering
	\caption{Accuracies of different methods.}
		\begin{tabular}{cccccccccc} 
			\toprule
			\multirow{1}{*} Dir &\multirow{3}{*}{Method} &\multicolumn{4}{c}{Dayton (64$\times$64)} &\multicolumn{4}{c}{Dayton (256$\times$256)} \\ \cmidrule(lr){3-6} \cmidrule(lr){7-10}
			\multirow{2}{*}{$\rightleftharpoons$}&  &\multicolumn{2}{c}{Top-1} &\multicolumn{2}{c}{Top-5} &\multicolumn{2}{c}{Top-1} &\multicolumn{2}{c}{Top-5} \\
			&  &\multicolumn{2}{c}{Accuracy(\%)} &\multicolumn{2}{c}{Accuracy(\%)}
			&\multicolumn{2}{c}{Accuracy(\%)}
			&\multicolumn{2}{c}{Accuracy(\%)} \\
			\hline
			\multirow{5}{*}{a2g} & Pix2pix~\cite{2} & 7.90 & 15.33  &27.61  &39.07  &6.80 &9.15 &23.55 &27.00 \\
			& X-Fork~\cite{6} & 16.63  & 34.73  & 46.35  &  70.01  & 30.00  & 48.68  &  61.57  &  78.84   \\
			& X-Seq~\cite{6} & 4.83  & 5.56  & 19.55  &  24.96  & 30.16  & 49.85  &  62.59  &  80.70   \\
			& SelectionGAN~\cite{12} & 45.37  & 79.00  & 83.48  &  97.74  & 42.11  & 68.12  &   77.74  &   92.89   \\
			&Ours & \multicolumn{1}{l}{{\bf 47.61}} &  \multicolumn{1}{l}{{\bf 81.24}} & \multicolumn{1}{l}{{\bf 86.12}}  &  \multicolumn{1}{l}{{\bf 98.44}}  & \multicolumn{1}{l}{{\bf 45.07}}   & \multicolumn{1}{l}{{\bf 77.12}}  & \multicolumn{1}{l}{{\bf 80.04}}   &  \multicolumn{1}{l}{{\bf 94.54}}  \\
			
			\hline
			\multirow{5}{*}{g2a} & Pix2pix~\cite{2} & 1.65  & 2.24  &7.49  &12.68 &10.23 &16.02 &30.90 &40.49 \\
			& X-Fork~\cite{6} &4.00 & 16.41 & 15.42 & 35.82 & 10.54
			& 15.29  &  30.76  &  37.32   \\
			& X-Seq~\cite{6} & 1.55 & 2.99  & 6.27  &  8.96 &12.30  & 19.62  &  35.95  &  45.94   \\
			&SelectionGAN~\cite{12} & 14.12  & 51.81  & 39.45  &  74.70  &  20.66  & 33.70  &   51.01  &   63.03   \\
			&Ours & \multicolumn{1}{l}{{\bf 14.26}} &  \multicolumn{1}{l}{{\bf 52.17}} & \multicolumn{1}{l}{{\bf 52.55}}  &  \multicolumn{1}{l}{{\bf 78.72}}  & \multicolumn{1}{l}{{\bf 20.81}}   & \multicolumn{1}{l}{{\bf 38.41}}  & \multicolumn{1}{l}{{\bf 55.51}}   &  \multicolumn{1}{l}{{\bf 65.84}}  \\
			\hline\\
	\end{tabular}
	\label{tb1:table1}
\end{table}

	\begin{table}[!t]
		\centering
		\caption{SSIM, PSNR, and KL score of different methods.}
			\begin{tabular}
				{cccccccc}
				\toprule
				\multirow{1}{*} Dir &\multirow{2}{*}{Method} &\multicolumn{3}{c}{Dayton(64$\times$64)} &\multicolumn{3}{c}{Dayton(256$\times$256)}  \\ \cmidrule(lr){3-5} \cmidrule(lr){6-8}
				
				\multirow{1}{*}{$\rightleftharpoons$}& &\multicolumn{1}{c}{SSIM} &\multicolumn{1}{c}{PSNR} &\multicolumn{1}{c}{KL} &\multicolumn{1}{c}{SSIM} &\multicolumn{1}{c}{PSNR} &\multicolumn{1}{c}{KL}\\
				\hline
				\multirow{5}{*}{a2g} & Pix2pix~\cite{2} & 0.4808  &19.4919  & 6.29$\pm$0.80  &0.4180  &17.6291 &38.26$\pm$1.88 \\
				& X-Fork~\cite{6} & 0.4921  & 19.6273  & 3.42$\pm$0.72 &   0.4963  &  19.8928  &  6.00$\pm$1.28   \\
				
				& X-Seq~\cite{6} & 0.5171  & 20.1049  &  6.22$\pm$0.87  &   0.5031  &  20.2803  &  5.93$\pm$1.32   \\
				& SelectionGAN~\cite{12} & 0.6865  & 24.6143  & 1.70$\pm$0.45   & 0.5938  &  23.8874  & 2.74 $\pm$0.86  \\
				&Ours &  \multicolumn{1}{l}{{\bf 0.7100}} & \multicolumn{1}{l}{{\bf 24.9674}}  &  \multicolumn{1}{l}{{\bf 1.55$\pm$0.51}}   & \multicolumn{1}{l}{{\bf 0.6524}}   & \multicolumn{1}{l}{{\bf 24.4012}}  & \multicolumn{1}{l}{{\bf 2.47$\pm$0.76}}     \\
				
				\hline
				\multirow{5}{*}{g2a} & Pix2pix~\cite{2} & 0.3675  &20.5135  & 6.39$\pm$0.90  &0.2693  &20.2177 &$7.88\pm$1.24 \\
				& X-Fork~\cite{6} & 0.3682  & 20.6933  & 4.55$\pm$0.84   & 0.2763  &  20.5978  &  6.92$\pm$1.15   \\
				
				& X-Seq~\cite{6} & 0.3663  & 20.4239  &  7.20$\pm$0.92   &  0.2725  &  20.2925  & 7.07$\pm$1.19   \\
				&SelectionGAN~\cite{12} & 0.5118  & 23.2657  & 2.25$\pm$0.56  & 0.3284  &  21.8066  &  3.55$\pm$0.87  \\
				
				&Ours &  \multicolumn{1}{l}{{\bf 0.6116}} & \multicolumn{1}{l}{{\bf 24.5445}}  &  \multicolumn{1}{l}{{\bf 2.13$\pm$0.48}} & \multicolumn{1}{l}{{\bf 0.3924}}   & \multicolumn{1}{l}{{\bf 22.7143}}  & \multicolumn{1}{l}{{\bf 3.17$\pm$0.82}}     \\
				\bottomrule\\
		\end{tabular}
		\label{tb2:table2}
	\end{table}
	
\begin{figure}[!t]
	\centerline{\includegraphics[width=0.8\linewidth]{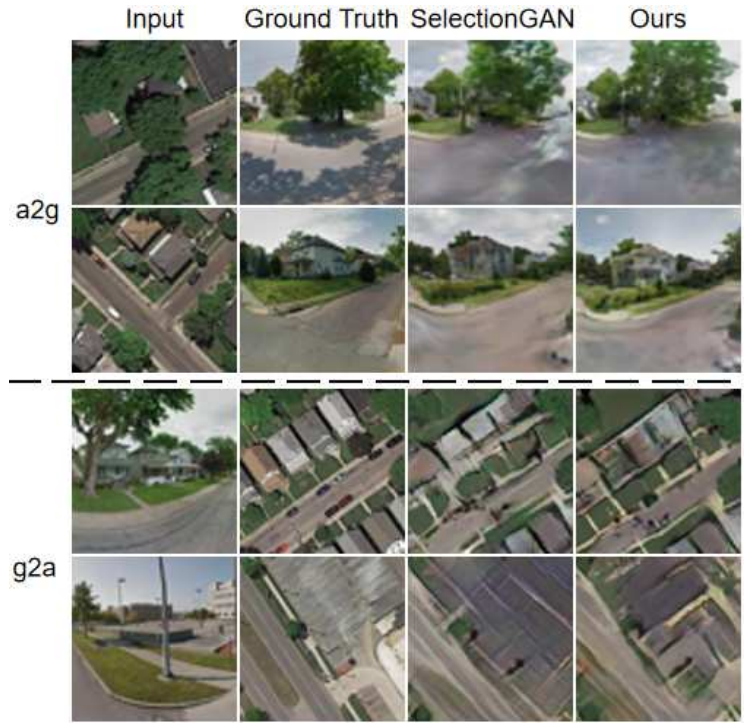}}
	\caption{Results generated by the proposed method and SelectionGAN~\cite{12} in $64 {\times} 64$ resolution in both a2g (top) and g2a (bottom) directions on the Dayton dataset.}
	\label{fig1}
\end{figure}
	
\noindent \textbf{Parameter Settings.} 
Similar to~\cite{12}, the low resolution ($64 {\times} 64$) experiments on the Dayton dataset are carried out for 100 epochs with batch size of 16, whereas the high resolution ($256 {\times} 256$) experiments for this dataset are trained for 35 epochs with batch size of 4. We also set  $\lambda_{1}{=}100$, $\lambda_{2}{=}1$, $\lambda_{3}{=}200$, $\lambda_{4}{=}2$ and $\lambda_{tv}{=}1e-6$ in Eq. \eqref{eq:loss}, and $\lambda$=4 in Eq. \eqref{eq:adv_loss}.
	
\noindent \textbf{Evaluation Protocol.}
We employ KL Score and top-k prediction accuracy as the evaluation metrics.
These metrics evaluate the generated images from a high-level feature space. We also employ pixel-level similarity metrics to evaluate our method,i.e., Structural-Similarity (SSIM) and Peak Signal-to-Noise Ratio (PSNR).
	
\noindent \textbf{State-of-the-art Comparisons.}
We compare the proposed model with exiting cross-view image translation methods, i.e., Pix2Pix~\cite{2}, X-Fork~\cite{6}, X-Seq~\cite{6} and SelectionGAN~\cite{12}.
Quantitative results of different metrics are shown in Tables~\ref{tb1:table1} and \ref{tb2:table2}.

We compute top-1 and top-5 accuracies in Table \ref{tb1:table1}.
As we can see, for lower resolution images ($64 {\times} 64$) our method outperforms the existing leading cross-view image translation methods.
For higher resolution images ($256 {\times} 256$), our method also achieves the best results on top-1 and top-5 accuracies. This shows the effectiveness of our method and the necessity of the proposed modules.

Moreover, we provide results of SSIM, PSNR, and KL scores Table \ref{tb2:table2}.
We observe that the proposed method is consistently superior to other leading methods, validating the effectiveness of the proposed method.

\begin{figure}[!t]
		\centerline{\includegraphics[width=0.8\linewidth]{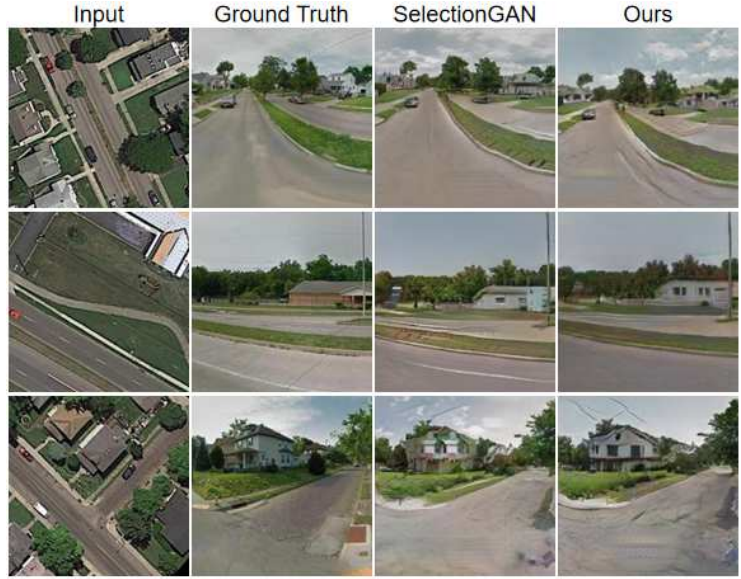}}
		\caption{Results generated by the proposed method and SelectionGAN~\cite{12} in $256 {\times} 256$ resolution in a2g direction on the Dayton dataset.}
		\label{fig5}
		\end{figure}

\begin{figure}[!t]
	\centerline{
		\includegraphics[width=0.8\linewidth]{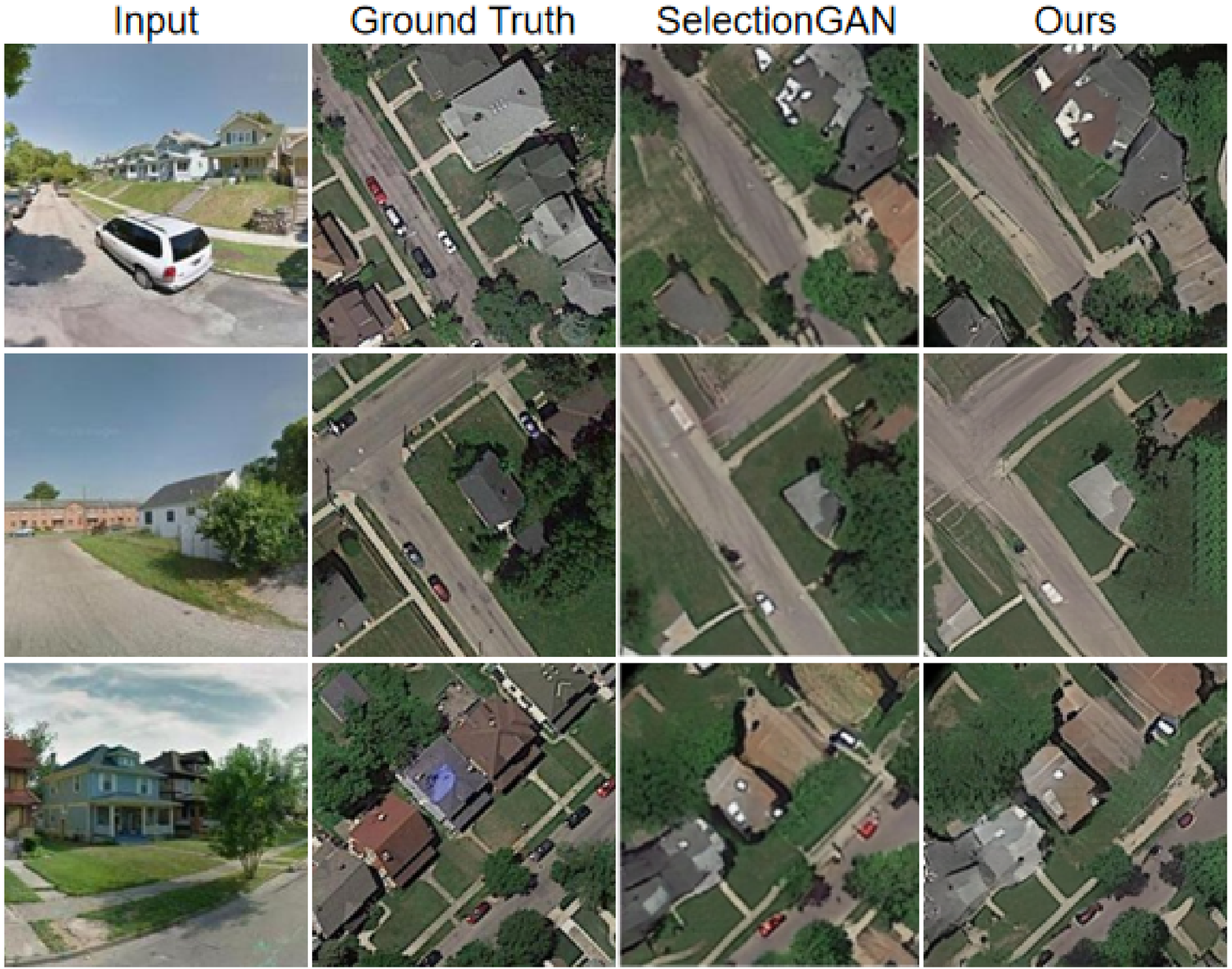}}
	\caption{Results generated by the proposed method and SelectionGAN~\cite{12} in $256 {\times} 256$ resolution in g2a direction on the Dayton dataset.}
	\label{fig6}
\end{figure}

\noindent \textbf{Qualitative Evaluation.} Qualitative results compared with the most related work, i.e., SelectionGAN~\cite{12} are shown in Fig.~\ref{fig5} and \ref{fig6}.
We can see that our method generates sharper details than SelectionGAN on objects/scenes, e.g., houses, buildings, roads, clouds, and cars.
For example, we can see that the houses generated by our method are more natural than those generated by SelectionGAN as shown in Fig. \ref{fig5}.

\begin{table}[!t]
	\centering
	\caption{Ablations study of the proposed method.}
		\begin{tabular}{cccc}
			\toprule
			Baseline & Method & PSNR & SSIM\\
			\midrule
			A & SGAN~\cite{12}            & 23.9310 & 0.6176  \\
			B & SGAN + AM       & 24.0539 & 0.6309  \\
			C & SGAN + AM + DC    & 24.3345 & 0.6507  \\
			D & SGAN + AM + DC + LS & {\bf 24.6421} & {\bf 0.6927}  \\
			\bottomrule
	\end{tabular}
	\label{tb3:table3}
\end{table}

\noindent \textbf{Ablation Study.}
We also conduct an ablation study in a2g (aerial-to-ground) direction on the Dayton dataset. To reduce the training time, we follow SelectionGAN and randomly select 1/3 samples from the whole 55,000/21,048 samples, i.e., around 18,334 samples for training and 7,017 samples for testing.
The proposed model consists of 4 baselines (A, B, C, D) as shown in Table~\ref{tb3:table3}.
Baseline A uses SelectionGAN (SGAN). Baseline B combines SGAN and the proposed attention mechanism (AM). Baseline C employs deformable convolution (DC) on baseline B.
Baseline D adopts the proposed loss function (LS).
It is obvious that as each module is added, we can obtain better results of both SSIM and PSNR metrics.
This means by adding the proposed attention mechanism, deformable convolution, and the proposed loss function, the overall performance can be further boosted.

\section{Conclusion}
In this paper, we propose a novel generative adversarial network based on deformable convolution and attention mechanisms for solving the challenging cross-view image generation task. We propose a novel attention mechanism to refine the feature maps, thus improving the ability of feature representation. We also embed deformed convolution in our generator to improve the network's ability for extracting object features at different scales.
Moreover, a novel semantic-guide adversarial loss is proposed to improve the whole network training, thus achieving a more robust and stable optimization.
Extensive experimental results show that the proposed method obtains better results than  state-of-the-art methods.

	%
	
	%
	%
	%

\begin{thebibliography}{8}
		\bibitem{1}
		Vo, N.N., Hays, J.: Localizing and orienting street views using overhead imagery. In: Proceedings of the European Conference on Computer Vision, pp. 494--509 (2016)

		\bibitem{2}
		Isola, P., Zhu, J.Y., Zhou, T., Efros, A.A.: Image-to-image translation with conditional adversarial networks. In: Proceedings of the IEEE Conference on Computer Vision and Pattern Recognition, pp. 1125--1134 (2017)
		
		\bibitem{3}
		Krizhevsky, A., Sutskever, I., Hinton, G.E.: Imagenet classification with deep convolutional neural networks. In: Advances in Neural Information Processing Systems, pp. 1097--1105 (2012)
				
		\bibitem{4}
		Kim, T., Cha, M., Kim, H., Lee, J.K., Kim, J.: Learning to discover cross-domain relations with generative adversarial networks. In: Proceedings of the 34th International Conference on Machine Learning-Volume 70, pp. 1857--1865 (2017)
		
		\bibitem{5}
		Zhu, X., Yin, Z., Shi, J., Li, H., Lin, D.: Generative adversarial frontal view to bird view synthesis. In: International Conference on 3D Vision, pp. 454--463 (2018)

		
		\bibitem{6}
    	Regmi, K., Borji, A.: Cross-view image synthesis using conditional gans. In: Proceedings of the IEEE Conference on Computer Vision and Pattern Recognition, pp. 3501--3510 (2018)
		
		\bibitem{7}
		Mathieu, M., Couprie, C., LeCun, Y.: Deep multi-scale video prediction beyond mean square error. In: International Conference on Learning Representations. (2016)
		
		\bibitem{8}
		Huang, R., Zhang, S., Li, T., He, R.: Beyond face rotation: Global and local perception gan for photorealistic and identity preserving frontal view synthesis. In: Proceedings of the IEEE International Conference on Computer Vision, pp. 2439--2448 (2017)

		
		\bibitem{9}
		Park, E., Yang, J., Yumer, E., Ceylan, D., Berg, A.C.: Transformation-grounded image generation network for novel 3d view synthesis. In: Proceedings of the IEEE Conference on Computer Vision and Pattern Recognition, pp. 3500--3509 (2017)
		
		\bibitem{10}
		Cordts, M., Omran, M., Ramos, S., Rehfeld, T., Enzweiler, M., Benenson, R., Franke, U., Roth, S., Schiele, B.: The cityscapes dataset for semantic urban scene understanding. In: Proceedings of the IEEE Conference on Computer Vision and Pattern Recognition, pp. 3213--3223 (2016)
		
		\bibitem{11}
		Yang, J., Reed, S.E., Yang, M.H., Lee, H.: Weakly-supervised disentangling with recurrent transformations for 3d view synthesis. In: Advances in Neural Information Processing Systems, pp. 1099--1107 (2015)
		
		\bibitem{12}
		Tang, H., Xu, D., Sebe, N., Wang, Y., Corso, J.J., Yan, Y.: Multi-channel attention selection gan with cascaded semantic guidance for cross-view image translation. In: Proceedings of the IEEE Conference on Computer Vision and Pattern Recognition, pp. 2417--2426 (2019)
		
		\bibitem{13}
		Woo, S., Park, J., Lee, J.Y., So~Kweon, I.: Cbam: Convolutional block attention module. In: Proceedings of the European Conference on Computer Vision, pp. 3--19 (2018)
		
		\bibitem{14}
		Dosovitskiy, A., Springenberg, J.T., Tatarchenko, M., Brox, T.: Learning to generate chairs, tables and cars with convolutional networks. In: IEEE transactions on pattern analysis and machine intelligence  \textbf{39}(4), 692--705 (2016)
		
		\bibitem{15}
	   Tatarchenko, M., Dosovitskiy, A., Brox, T.: Multi-view 3d models from single images with a convolutional network. In: Proceedings of the European Conference on Computer Vision, pp. 322--337 (2016)
		
		\bibitem{16}
		Zhou, T., Tulsiani, S., Sun, W., Malik, J., Efros, A.A.: View synthesis by appearance flow. In: Proceedings of the European Conference on Computer Vision, pp. 286--301 (2016)
		
		\bibitem{17}
        Zhai, M., Bessinger, Z., Workman, S., Jacobs, N.: Predicting ground-level scene layout from aerial imagery. In: Proceedings of the IEEE Conference on Computer Vision and Pattern Recognition, pp. 867--875 (2017)
		
		\bibitem{18}
		Lin, T.Y., Belongie, S., Hays, J.: Cross-view image geolocalization. In: Proceedings of the IEEE Conference on Computer Vision and Pattern Recognition, pp. 891--898 (2013)
		
		\bibitem{19}
		Nelson, B.K., Cai, X., Nebenf{\"u}hr, A.: A multicolored set of in vivo organelle markers for co-localization studies in arabidopsis and other plants. The Plant Journal, 51(6), pp. 1126--1136 (2007)
		
		\bibitem{20}
		Lin, T.Y., Cui, Y., Belongie, S., Hays, J.: Learning deep representations for ground-to-aerial geolocalization. In: Proceedings of the IEEE Conference on Computer Vision and Pattern Recognition, pp. 5007--5015 (2015)
		
		\bibitem{21}
		Workman, S., Souvenir, R., Jacobs, N.: Wide-area image geolocalization with aerial reference imagery. In: Proceedings of the IEEE International Conference on Computer Vision, pp. 3961--3969 (2015)
		
		\bibitem{22}
		Hinton, G.E., Osindero, S., Teh, Y.W.: A fast learning algorithm for deep belief nets. Neural computation, 18(7), pp. 1527--1554 (2006)
		
		\bibitem{23}
		Hinton, G.E., Salakhutdinov, R.R.: A better way to pretrain deep boltzmann machines. In: Advances in Neural Information Processing Systems, pp. 2447--2455 (2012)
		
		\bibitem{24}
        Goodfellow, I., Pouget-Abadie, J., Mirza, M., Xu, B., Warde-Farley, D., Ozair, S., Courville, A., Bengio, Y.: Generative adversarial nets. In: Advances in Neural Information Processing Systems, pp. 2672--2680 (2014)		
        
		\bibitem{25}
		Wang, X., Gupta, A.: Generative image modeling using style and structure adversarial networks. In: Proceedings of the European Conference on Computer Vision, pp. 318--335 (2016)
		
		\bibitem{26}
		Karras, T., Aila, T., Laine, S., Lehtinen, J.: Progressive growing of gans for improved quality, stability, and variation. In: International Conference on Learning Representations. (2017)
		
		\bibitem{27}
		Gulrajani, I., Ahmed, F., Arjovsky, M., Dumoulin, V., Courville, A.C.: Improved training of wasserstein gans. In: Advances in Neural Information Processing Systems, pp. 5767--5777 (2017)
		
        \bibitem{28}
        Tang, H., Wang, W., Xu, D., Yan, Y. and Sebe, N.: Gesturegan for hand gesture-to-gesture translation in the wild. In: Proceedings of the 26th ACM international conference on Multimedia, pp. 774-782 (2018).
		
		\bibitem{29}
		Zhang, H., Goodfellow, I., Metaxas, D., Odena, A.: Self-attention generative adversarial networks. In: International Conference on Machine Learning, pp. 7354-7363,  (2019)
		
		\bibitem{30}
		Radford, A., Metz, L., Chintala, S.: Unsupervised representation learning with deep convolutional generative adversarial networks. In: International Conference on Learning Representations. (2016)
		
		\bibitem{31}
		Liu, G., Tang, H., Latapie, H. and Yan, Y.: Exocentric to Egocentric Image Generation Via Parallel Generative Adversarial Network. In: IEEE International Conference on Acoustics, Speech and Signal Processing, pp. 1843-1847 (2020)
		
		\bibitem{32}
		Dai, J., Qi, H., Xiong, Y., Li, Y., Zhang, G., Hu, H., Wei, Y.: Deformable convolutional networks. In: Proceedings of the IEEE International Conference on Computer Vision, pp. 764--773 (2017)
		
		\bibitem{33}
		Tang, H., Xu, D., Yan, Y., Torr, P.H. and Sebe, N.: Local class-specific and global image-level generative adversarial networks for semantic-guided scene generation. In: Proceedings of the IEEE Conference on Computer Vision and Pattern Recognition, pp. 7870-7879 (2020)
		
	
	\end{thebibliography}

\end{document}